\documentclass{article}

\usepackage[final]{neurips_2023}
\makeatletter
\renewcommand{\@noticestring}{}
\makeatother

\usepackage[utf8]{inputenc}
\usepackage[T1]{fontenc}
\usepackage{hyperref}
\usepackage{url}
\usepackage{booktabs}
\usepackage{amsfonts}
\usepackage{amsmath}
\usepackage{nicefrac}
\usepackage{microtype}
\usepackage{xcolor}
\usepackage{graphicx}
\usepackage{algorithm}
\usepackage{algorithmic}
\usepackage{enumitem}
\usepackage{tikz}
\usetikzlibrary{arrows.meta, positioning, shapes.geometric, calc, decorations.pathreplacing, shadows}
\usepackage{pgfplots}
\pgfplotsset{compat=1.17}
\usepgfplotslibrary{groupplots, fillbetween}
\usepackage{listings}
\usepackage{multirow}
\usepackage{pifont}

\newcommand{\xmark}{\ding{55}}
\newcommand{\auto}{\textsc{Auto}}
\newcommand{\autobench}{\textsc{Auto-Bench}}

\definecolor{recblue}{RGB}{41,98,255}
\definecolor{gateorange}{RGB}{230,126,34}
\definecolor{runnergreen}{RGB}{39,174,96}
\definecolor{failred}{RGB}{192,57,43}
\definecolor{irpurple}{RGB}{125,60,255}

\title{\auto: The AGI Compiler}

\author{
  Jaber Jaber\thanks{Correspondence: \texttt{jaber@rightnowai.co}} \\
  RightNow AI\\
  \texttt{jaber@rightnowai.co} \\
  \And
  Osama Jaber \\
  RightNow AI\\
  \texttt{osama@rightnowai.co} \\
}

\begin{document}
\maketitle

\begin{center}
\includegraphics[height=1.1cm]{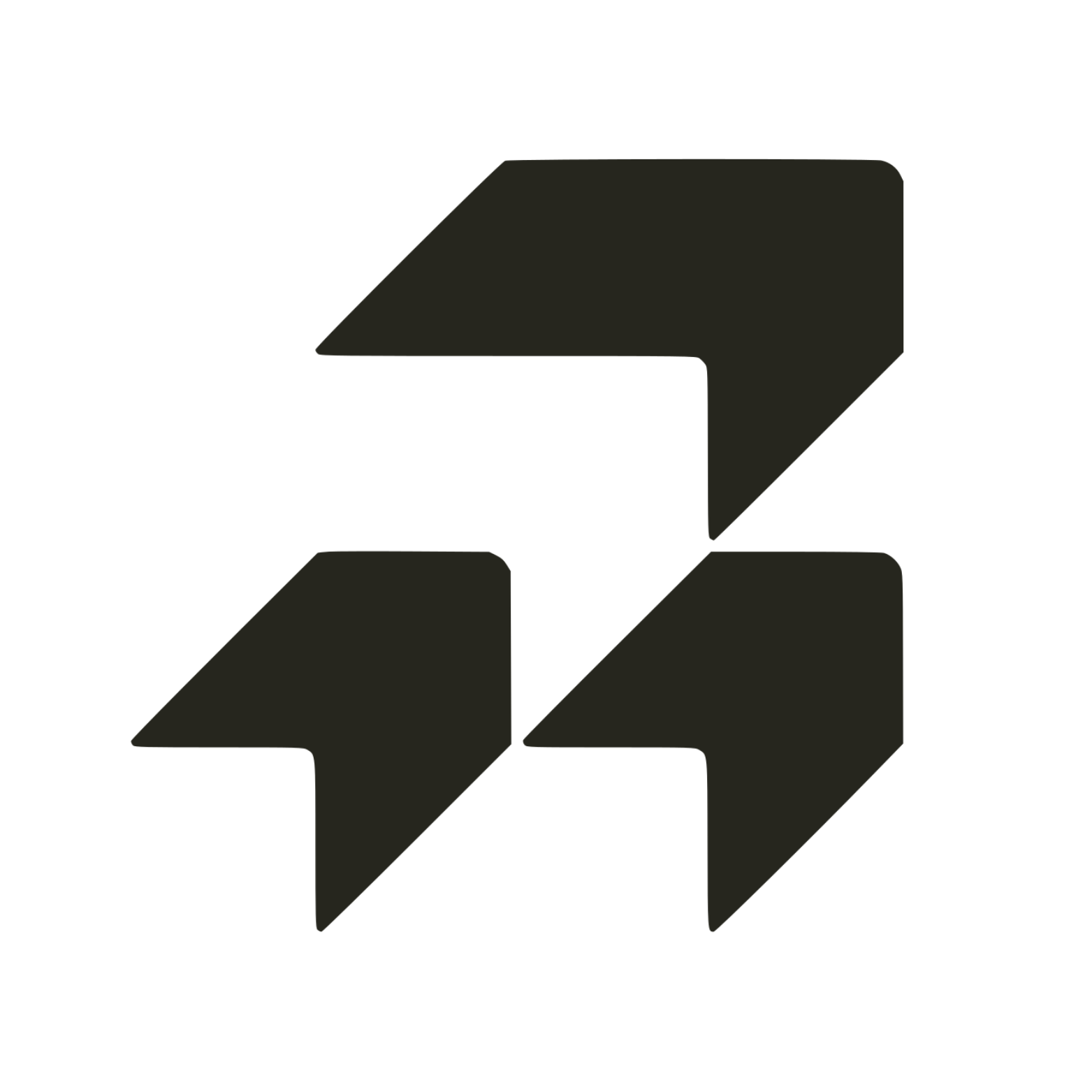}
\end{center}
\vspace{-0.3em}

\begin{abstract}
Every LLM agent run re-derives its behavior token by token on a frontier model: brilliant, expensive, slow, and unbounded. We present \auto, a compiler that records live agent behavior, measures which parts are secretly deterministic, extracts them into verified programs or distilled specialists, and emits \emph{cognition binaries}: WebAssembly artifacts whose manifests carry measured guarantees and whose declared capabilities are physically enforced by the sandbox. A tiered runtime executes compiled behavior behind conformally calibrated guards; guard trips deopt to the reference agent, and the captured trace recompiles back down, so nothing is figured out twice. We use ``AGI compiler'' in one narrow, testable sense: a system that autonomously converts novel experience into permanent, verified, near-free skill while measuring what it does not know. On \autobench, a benchmark we introduce and pre-register, 87.1\% of 560 recorded frontier-agent spans are witnessed-deterministic (three of the four censused task families measure 100.0\%). On a 300-item stream with three scheduled distribution shifts, the closed loop compiles three artifact generations and drives marginal cost from 59 to 2 micro-dollars per item (6.4$\times$ end-to-end) at 96.9\% parity on witnessed inputs with zero errors. The same stream also quantifies the failure modes: a loose guard silently mislabels 48.9\% of compiled answers, and an unfaithful deopt reference causes the verification gate to refuse recompilation. Calibration and reference fidelity, not model capability, decide whether cheap stays correct. Code: \url{https://github.com/RightNow-AI/auto}.
\end{abstract}

\section{Introduction}

An LLM agent that triages support tickets makes the same routing decision thousands of times a day, and pays a frontier model to re-derive it from scratch every single time. The tax is real money. We measure a single-call routing agent at 55.6 micro-dollars and 904\,ms per decision, and a three-call pipeline at 192.5 micro-dollars and 2.9 seconds per run. At production volume this is the biggest line item in agent deployment, and it buys nothing new. The model computes an answer it already computed yesterday, with no guarantee it computes the same one.

The history of computing is a sequence of compilation events: hand computation became written procedure, procedure became stored program, interpreted programs became compiled binaries. Agent cognition today sits where interpreted code sat: flexible, portable, and paying the interpreter on every execution. The missing artifact is the binary: a cheap, fast, reproducible, auditable executable form of one specific competence. Missing with it are the toolchain that produces it from running behavior and the runtime that falls back to the interpreter when the world changes.

Existing work reduces the tax without producing the artifact. Model routing~\citep{chen2023frugalgpt,ong2024routellm} chooses a cheaper model per query, and probabilistic cascades~\citep{dohan2022cascades} compose several; every query is still interpreted. Semantic caches~\citep{bang2023gptcache} replay stored answers without verification or capability confinement; they compute nothing new, only near-duplicates of past answers. Distillation~\citep{hinton2015distilling,hsieh2023distilling} produces smaller models offline, without a recording front end, a behavioral contract, or a fallback tier. Agent skill libraries~\citep{wang2023voyager} accumulate code the agent itself wrote, verified by nothing but the agent's own judgment and unconfined. None of these systems measure the property the whole enterprise depends on: how much of an agent's behavior is deterministic enough to compile at all.

\auto{} is that toolchain. It records live agent runs through drop-in SDK shims; measures the witnessed-deterministic fraction of behavior span by span; lowers recorded behavior to a typed task-graph IR whose nodes carry capability effects and uncertainty classes; extracts the symbolic parts by enumerative and LLM-guided counterexample-driven synthesis~\citep{solar2006combinatorial}; distills the residue into small specialists; verifies every candidate against a behavioral contract by differential replay; and emits a signed WebAssembly artifact~\citep{haas2017webassembly} whose manifest reports only measured numbers. A tiered runtime executes the artifact behind a conformally calibrated guard~\citep{angelopoulos2021gentle}; out-of-distribution inputs deopt to the reference agent, the resulting trace is captured, and a recompile folds the novelty into the next artifact generation. We call this loop the \emph{ratchet}.

The key insight is that verification, not synthesis, is the product. Most agent behavior is secretly a parser, and many methods can guess the program; the hard part is refusing the guesses that are wrong. \auto{} makes the behavioral contract the type system: a candidate that fails differential replay against recorded reality, at thresholds the contract declares before compilation, does not become an artifact. Every claim the system makes is measured, including its refusals. That is what lets a compiled artifact carry authority that a cache entry or a bare distilled model cannot.

\paragraph{Why ``the AGI compiler.''}
General intelligence, wherever its definition lands, has one non-negotiable property: it converts experience into skill. A system that re-derives the same decision at full price on every encounter does not accumulate competence; it rents it. Cognitive science names the missing operation proceduralization, deliberate cognition becoming automatic skill, and the classical architectures treated it as the centerpiece of general intelligence: SOAR compiled deliberation into productions~\citep{laird1987soar}, and explanation-based generalization turned single experiences into reusable rules~\citep{mitchell1986ebl}. Frontier agents have the deliberation and lack the compiler. \auto{} is that compiler: the mechanism by which an agent system's experience becomes permanent, verified, near-free skill, bounded by a calibrated measure of what it has not yet learned. The title is therefore a claim of necessity, not of achievement. We do not claim to have built a general intelligence; we claim that any system deserving the name must contain this loop, and we build the loop, close it live, and measure both its economics and its failure modes.

Our contributions:
\begin{enumerate}[leftmargin=1.4em,itemsep=0.1em]
\item A recording-to-binary compilation pipeline for LLM agent behavior, with a typed effect-carrying IR, synthesis and distillation passes, and contract-gated emission (\S\ref{sec:design}).
\item Physical capability confinement: pure artifacts compile to WebAssembly with \emph{zero} imports, and tool-using artifacts import exactly one audited host function, so an artifact cannot exceed its declared capabilities (\S\ref{sec:verification}).
\item A tiered runtime whose guards are conformally calibrated abstention mechanisms, with deopt, trace capture, and autonomous recompilation, embeddable in-process down to 18.2\,$\mu$s per call (\S\ref{sec:design}).
\item Judged verification: an LLM judge, itself spend-capped and ledgered, arbitrates semantic equivalence for generative spans, with declared agreement thresholds deciding acceptance (\S\ref{sec:verification}).
\item \autobench{}, a pre-registered benchmark measuring what no capability evaluation measures: the determinism census, parity-gated compression, calibrated ignorance, and the ratchet curve, cost per thought as a function of experience (\S\ref{sec:eval}).
\item Measured failure modes with equal weight: 48.9\% silent wrongness under loose calibration, gate-refused recompilation under unfaithful references, and honest refusal of the generative residue at scale (\S\ref{sec:eval}).
\end{enumerate}

\begin{figure}[t]
\centering
\begin{tikzpicture}
\begin{axis}[
  width=12.5cm, height=6.4cm,
  xmin=0, xmax=300, ymin=0, ymax=18500,
  xtick={0,50,100,150,200,250,300},
  ytick={0,5000,10000,15000},
  yticklabels={0, {5{,}000}, {10{,}000}, {15{,}000}},
  scaled y ticks=false,
  xlabel={tasks completed}, ylabel={total spent ($\mu$\$)},
  grid=major, grid style={black!8},
  tick label style={font=\scriptsize},
  label style={font=\footnotesize},
  clip=false,
]
\addplot[failred, line width=1.6pt] coordinates {(0,0) (300,17692)};
\node[font=\scriptsize\bfseries\sffamily, text=failred, anchor=south east, rotate=15]
  at (axis cs:262,16050) {paying the model for every task};
\node[font=\scriptsize\bfseries\sffamily, text=failred, anchor=west]
  at (axis cs:301,17692) {17{,}692 $\mu$\$};
\addplot[recblue, line width=1.8pt] coordinates {
(0,0) (1,63) (2,120) (3,181) (4,239) (5,296) (6,353) (7,413) (8,472) (10,472) (20,472) (30,472) (40,472) (50,535) (57,595) (60,595) (61,655) (62,718) (68,776) (70,837) (71,897) (72,957) (74,1017) (76,1077) (77,1135) (78,1198) (80,1257) (83,1314) (84,1374) (90,1374) (94,1434) (100,1434) (110,1434) (120,1490) (121,1550) (124,1606) (130,1606) (132,1666) (134,1726) (139,1786) (140,1786) (141,1841) (143,1900) (148,1958) (150,1958) (151,2015) (154,2070) (160,2070) (170,2070) (180,2070) (189,2127) (190,2127) (200,2127) (203,2187) (210,2187) (216,2246) (220,2246) (224,2306) (227,2365) (229,2425) (230,2482) (233,2541) (237,2601) (240,2601) (247,2658) (250,2658) (260,2658) (269,2715) (270,2715) (280,2715) (290,2715) (293,2775) (300,2775)
};
\node[font=\scriptsize\bfseries\sffamily, text=recblue, anchor=west]
  at (axis cs:301,2775) {2{,}775 $\mu$\$};
\node[font=\scriptsize\bfseries\sffamily, text=recblue, anchor=west]
  at (axis cs:88,4100) {\auto: same tasks, same verified answers};
\node[font=\scriptsize\sffamily, text=recblue!70!black, anchor=west] (flat)
  at (axis cs:8,6600) {flat = compiled: tasks cost nothing};
\draw[recblue!70!black, thick, -{Stealth[length=3.5pt]}] ($(flat.south west)+(0.35,0)$) -- (axis cs:30,750);
\draw[{Stealth[length=5pt]}-{Stealth[length=5pt]}, runnergreen!55!black, line width=1.4pt]
  (axis cs:228,3200) -- (axis cs:228,13100);
\node[font=\footnotesize\bfseries\sffamily, text=runnergreen!40!black, anchor=west, align=left]
  at (axis cs:233,8100) {$6.4\times$ less spent\\[-2pt]{\scriptsize on the same 300 tasks}};
\draw[gateorange, line width=1.4pt] (axis cs:50,0) -- (axis cs:50,900);
\draw[gateorange, line width=1.4pt] (axis cs:120,0) -- (axis cs:120,900);
\draw[gateorange, line width=1.4pt] (axis cs:200,0) -- (axis cs:200,900);
\node[font=\tiny\sffamily, text=gateorange!85!black, anchor=west] at (axis cs:206,800)
  {$\leftarrow$ task mix changes (paid once, then compiled)};
\end{axis}
\end{tikzpicture}
\caption{The central result in its simplest form: cumulative spend over the same 300 live tasks. The red line pays the frontier model for every task. The blue line is \auto, measured: a brief paid burst while it learns, then flat stretches where compiled execution costs nothing, with small rises where the task mix changes and novelty is paid for once. Same tasks, answers verified against the same agent, $6.4\times$ less money. Figure~\ref{fig:ratchet} gives the per-task detail.}
\label{fig:cumulative}
\end{figure}
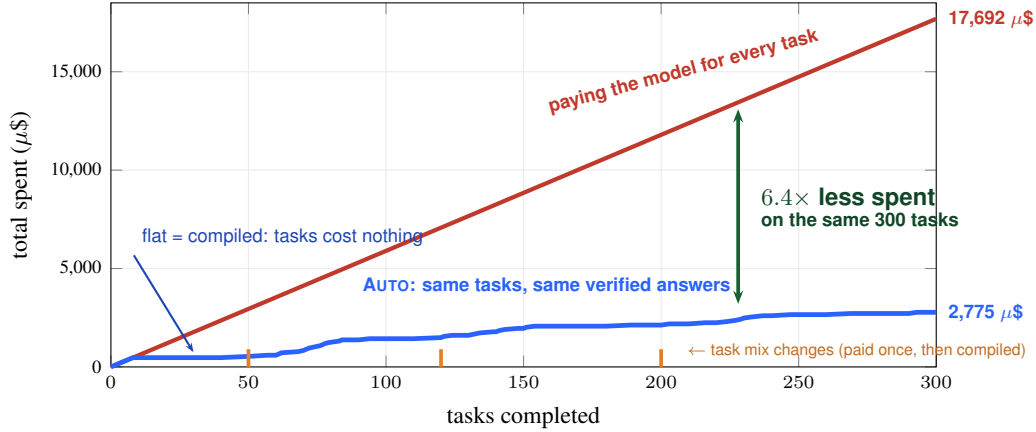

Figure~\ref{fig:cumulative} previews the central result in its simplest form: the same 300 tasks, priced twice.

Section~\ref{sec:related} situates \auto{} among systems that reduce the interpretation tax without producing the artifact, and \S\ref{sec:limitations} states what the current evidence does and does not support.

\section{Related Work}
\label{sec:related}

\paragraph{Cost reduction for LLM serving.} FrugalGPT~\citep{chen2023frugalgpt} cascades queries through cheaper models; RouteLLM~\citep{ong2024routellm} learns routing from preference data; cascades~\citep{dohan2022cascades} formalize multi-model composition. All reduce the per-query price of interpretation; none produce an executable artifact, a behavioral guarantee, or a confinement boundary. Speculative decoding~\citep{leviathan2023fast} accelerates the decoder itself and is complementary. GPTCache~\citep{bang2023gptcache} replays semantically similar answers, a memoization of outputs; \auto{} compiles the function, verifies it against witnessed behavior, and abstains outside calibration rather than serving the nearest hit.

\paragraph{Agents, tools, and skills.} ReAct~\citep{yao2022react}, Toolformer~\citep{schick2023toolformer}, and Gorilla~\citep{patil2023gorilla} established tool-calling agents; MemGPT~\citep{packer2023memgpt} manages agent memory; Voyager~\citep{wang2023voyager} accumulates a skill library of self-written code, the closest ancestor of our ratchet. Voyager's skills are unverified, unconfined, and curated by the agent itself; \auto's artifacts pass a differential gate, run in a capability-confined sandbox, and carry manifests of measured guarantees. Agent benchmarks~\citep{liu2023agentbench,jimenez2023swebench} measure task success of the interpreter; \autobench{} measures the compilability, calibration, and amortization of the interpreter's behavior.

\paragraph{Compilation and synthesis.} DSPy~\citep{khattab2023dspy} compiles declarative LM pipelines by optimizing prompts and weights, compilation \emph{into} better interpretation; the LLM Compiler~\citep{kim2023llmcompiler} parallelizes function-calling plans at runtime. \auto{} compiles \emph{out of} interpretation into programs that no longer call a model. Our extraction pass is counterexample-guided inductive synthesis~\citep{solar2006combinatorial} with an LLM proposer whose candidates are verified against recorded traces in a sandboxed evaluator; the frozen IR design follows the progressive-lowering discipline of MLIR~\citep{lattner2021mlir}, rebuilt around capability effects rather than tensors. Knowledge distillation~\citep{hinton2015distilling,hsieh2023distilling} supplies our residue pass, wrapped in a train/eval/accept loop whose acceptance is the same differential gate.

\paragraph{Older roots.} The ratchet is a modern instance of ideas the symbolic era named decades ago: SOAR's chunking converted deliberate problem solving into compiled productions~\citep{laird1987soar}; explanation-based generalization~\citep{mitchell1986ebl} turned single experiences into reusable rules; the knowledge-compilation literature~\citep{darwiche2002knowledge} studied offline conversion into tractable executable forms. \auto{} revives this program with a frontier model as the interpreter, WebAssembly~\citep{haas2017webassembly} as the target, and conformal prediction~\citep{angelopoulos2021gentle} as the boundary between compiled confidence and honest ignorance.

\section{System Design}
\label{sec:design}

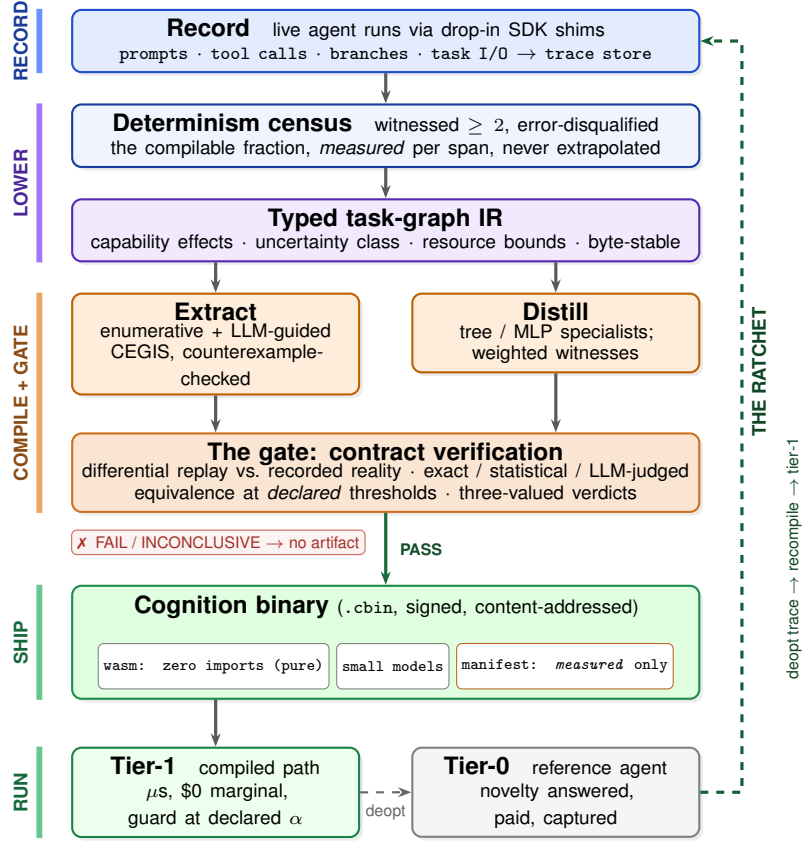
\begin{figure}[t]
\centering
\begin{tikzpicture}[
  font=\footnotesize\sffamily,
  box/.style={draw, rounded corners=3.5pt, thick, minimum height=0.6cm,
              align=center, inner sep=3.5pt,
              drop shadow={shadow xshift=1.4pt, shadow yshift=-1.4pt, opacity=0.15}},
  wide/.style={box, text width=8.05cm},
  half/.style={box, text width=3.55cm},
  chip/.style={draw, rounded corners=2pt, fill=white, inner sep=2pt, font=\tiny\ttfamily},
  arr/.style={-{Stealth[length=5pt]}, very thick, black!65},
  darr/.style={-{Stealth[length=5pt]}, thick, dashed},
  rail/.style={font=\scriptsize\bfseries\sffamily, rotate=90, anchor=south},
  node distance=0.4cm,
]
\node[wide, fill=recblue!13, draw=recblue!80!black] (rec)
  {\textbf{Record} \; {\scriptsize live agent runs via drop-in SDK shims}\\[-1.5pt]
   {\scriptsize\ttfamily prompts \(\cdot\) tool calls \(\cdot\) branches \(\cdot\) task I/O \(\to\) trace store}};
\node[wide, fill=recblue!7, draw=recblue!60!black, below=of rec] (census)
  {\textbf{Determinism census} \; {\scriptsize witnessed \(\geq 2\), error-disqualified}\\[-1.5pt]
   {\scriptsize the compilable fraction, \emph{measured} per span, never extrapolated}};
\node[wide, fill=irpurple!11, draw=irpurple!70!black, below=of census] (ir)
  {\textbf{Typed task-graph IR}\\[-1.5pt]
   {\scriptsize capability effects \(\cdot\) uncertainty class \(\cdot\) resource bounds \(\cdot\) byte-stable}};
\node[half, fill=orange!14, draw=orange!75!black, below=0.4cm of ir.south west, anchor=north west] (synth)
  {\textbf{Extract}\\[-1.5pt]{\scriptsize enumerative + LLM-guided\\[-2pt] CEGIS, counterexample-checked}};
\node[half, fill=orange!14, draw=orange!75!black, below=0.4cm of ir.south east, anchor=north east] (distill)
  {\textbf{Distill}\\[-1.5pt]{\scriptsize tree / MLP specialists;\\[-2pt] weighted witnesses}};
\node[wide, fill=gateorange!20, draw=gateorange!90!black, below=2.25cm of ir] (gate)
  {\textbf{The gate: contract verification}\\[-1.5pt]
   {\scriptsize differential replay vs.\ recorded reality \(\cdot\) exact / statistical / LLM-judged\\[-2pt]
    equivalence at \emph{declared} thresholds \(\cdot\) three-valued verdicts}};
\node[wide, fill=green!12, draw=runnergreen!80!black, below=0.95cm of gate] (cbin)
  {\textbf{Cognition binary} {\scriptsize(\texttt{.cbin}, signed, content-addressed)}\\[1.5pt]
   \begin{tikzpicture}[baseline]
     \node[chip, draw=black!50] (c1) {wasm: zero imports (pure)};
     \node[chip, draw=black!50, right=2.5pt of c1] (c2) {small models};
     \node[chip, draw=gateorange!80!black, right=2.5pt of c2] (c3) {manifest: \emph{measured} only};
   \end{tikzpicture}};
\node[half, minimum height=1.05cm, fill=green!7, draw=runnergreen!75!black, below=0.62cm of cbin.south west, anchor=north west] (t1)
  {\textbf{Tier-1} \; {\scriptsize compiled path}\\[-1.5pt]{\scriptsize \(\mu\)s, \$0 marginal, guard at declared \(\alpha\)}};
\node[half, minimum height=1.05cm, fill=black!4, draw=black!50, below=0.62cm of cbin.south east, anchor=north east] (t0)
  {\textbf{Tier-0} \; {\scriptsize reference agent}\\[-1.5pt]{\scriptsize novelty answered, paid, captured}};
\draw[arr] (rec) -- (census);
\draw[arr] (census) -- (ir);
\draw[arr] (ir.south -| synth) -- (synth.north);
\draw[arr] (ir.south -| distill) -- (distill.north);
\draw[arr] (synth.south) -- (gate.north -| synth.south);
\draw[arr] (distill.south) -- (gate.north -| distill.south);
\draw[arr, runnergreen!60!black] (gate.south) -- node[right=1.5pt, font=\tiny\bfseries\sffamily, text=runnergreen!45!black]{PASS} (cbin.north);
\node[draw=failred!80, rounded corners=2pt, fill=failred!6, font=\tiny\sffamily, text=failred!85!black,
      inner sep=2.2pt, anchor=north west] (fail) at ($(gate.south west)+(0.0,-0.22)$)
      {\xmark\; FAIL / INCONCLUSIVE \(\to\) no artifact};
\draw[arr] (cbin.south -| t1) -- (t1.north);
\draw[darr, black!60] (t1.east) -- node[below=1pt, font=\tiny\sffamily]{deopt} (t0.west);
\draw[darr, runnergreen!55!black, very thick]
  (t0.east) -- ++(0.55,0) |- (rec.east)
  node[font=\scriptsize\bfseries\sffamily, text=runnergreen!40!black, rotate=90, anchor=north, pos=0.28]{THE RATCHET};
\node[font=\tiny\sffamily, text=runnergreen!40!black, rotate=90, anchor=north]
  at ($(t0.east)+(1.02,3.1)$) {deopt trace \(\to\) recompile \(\to\) tier-1};
\foreach \a/\b/\c/\t in {rec/rec/recblue/RECORD, census/ir/irpurple/LOWER,
                          synth/gate/gateorange/COMPILE + GATE, cbin/cbin/runnergreen/SHIP, t1/t1/runnergreen/RUN}{
  \draw[\c!70, line width=2.2pt]
    ($(\a.north west)+(-0.42,0)$) -- ($(\b.south west)+(-0.42,0)$)
    node[rail, midway, text=\c!60!black]{\t};
}
\end{tikzpicture}
\caption{The \auto{} pipeline. Recorded behavior is measured, lowered to an effect-typed IR, extracted or distilled, and emitted only through the verification gate; a failing or unmeasurable contract never emits (red). The runtime splits execution across tiers behind a conformally calibrated guard, and the ratchet (green) recompiles every captured deopt, so nothing is figured out twice.}
\label{fig:architecture}
\end{figure}

\auto{} is a Rust workspace of 16 crates (36{,}843 lines of Rust across 96 files, 32 numbered architecture decision records), organized as a classical compiler: front end, IR, passes, back end, runtime, registry. Figure~\ref{fig:architecture} shows the pipeline; Algorithm~\ref{alg:ratchet} gives the runtime loop.

\paragraph{Recording and the census.} Drop-in tracer shims record every effectful span of a live agent: model calls, tool calls, branch decisions, environment reads, and whole-task I/O, with the paid call executing inside the traced closure so latency and provider-reported cost attach to the span. The determinism report then makes the thesis measurable before any compilation: a span signature is \emph{witnessed-deterministic} when observed at least twice with identical output and no errors, and fractions cover witnessed spans only, never extrapolated.

\paragraph{IR and passes.} Recorded behavior lowers to a typed task graph whose nodes carry capability effects, memory effects, an uncertainty class (\texttt{Deterministic}\,$\mid$\,\texttt{Probabilistic}\,$\mid$\,\texttt{Generative}), and resource bounds; serialization is byte-stable, a property-tested invariant. Extraction runs enumerative search over a closed 17-operation DSL and, when that refuses, LLM-guided CEGIS: a frontier model proposes candidate programs, a sandboxed evaluator executes them against every witnessed input, and counterexamples drive the next round. The checker is the authority: in our evaluation it catches a proposer attempting a constant-output shortcut and kills it with a counterexample (\S\ref{sec:eval}). Multi-span regions compile too: each stage and each inter-stage glue edge becomes its own synthesis problem, and tool-call stages become declared capability boundaries. The residue distills into decision trees or MLPs, optionally trained on \emph{every} witnessed output weighted by witness count rather than a single majority pick.

\paragraph{Runtime and embeddings.} A guard evaluates every compiled entry: lexical trigram sketches (Jaccard or hashed-cosine) over witnessed inputs, with the abstention threshold set by split-conformal calibration over leave-one-out scores at a declared miscoverage $\alpha$. In-distribution inputs run tier-1; trips deopt to tier-0 and the answer is conformance-checked, served, and ingested for recompilation. The runtime embeds anywhere along a measured latency ladder: 736\,ms median for the frontier reference call, 21\,ms behind an HTTP serving boundary, 290\,$\mu$s as a resident stdio process, 54.1\,$\mu$s in-process in Python (pyo3), and 18.2\,$\mu$s in-process in Node (napi), the last two measuring the embedding boundary on a trivial artifact rather than model inference. Capability artifacts run embedded as well, their declared tools mapped to host functions under an exactly-declared rule with per-answer call budgets.

\begin{algorithm}[t]
\caption{The ratchet: tiered execution with autonomous recompilation}
\label{alg:ratchet}
\begin{algorithmic}[1]
\STATE \textbf{given} contract $C$ (declared thresholds), trace store $S$, artifact $A \gets \emptyset$, recompile stride $K$
\FOR{each input $x$ in the request stream}
  \IF{$A \neq \emptyset$ \AND $\mathrm{guard}_A(x) \leq \tau_A$}
     \STATE serve $A(x)$ \COMMENT{tier-1: marginal cost $\approx 0$}
  \ELSE
     \STATE $y \gets$ reference agent$(x)$; serve $y$; ingest $(x,y)$ into $S$ \COMMENT{tier-0: paid, captured}
  \ENDIF
  \IF{$K$ new distinct inputs ingested since last compile}
     \STATE $A' \gets \mathrm{compile}(S, C)$ \COMMENT{extract/distill, then the gate}
     \IF{$A'$ passes $C$} \STATE $A \gets A'$ \COMMENT{hot-swap; guard recalibrated on new witnesses}
     \ELSE \STATE keep $A$; log the refusal \COMMENT{a failing contract never emits}
     \ENDIF
  \ENDIF
\ENDFOR
\end{algorithmic}
\end{algorithm}

\section{Verification as the Type System}
\label{sec:verification}

A contract declares examples, properties, latency and cost budgets, and acceptance thresholds; its content hash is the artifact's type. Three rules make the gate trustworthy. First, three-valued verdicts: a check that cannot be measured is \textsc{Inconclusive}, never rounded up to \textsc{Pass}, and both \textsc{Fail} and \textsc{Inconclusive} block emission. Second, differential replay is the ground truth: every distinct recorded input replays through the candidate, byte-equal by default; a contract may instead declare a statistical agreement threshold, and, for generative spans, \emph{judged} equivalence, where an LLM judge (itself spend-capped and ledgered) arbitrates byte-divergent outputs and the declared threshold still decides. A judged example with no judge supplied is \textsc{Inconclusive}, never silently exact. Third, manifests report measured numbers or null, each parity claim carrying the content-addressed identifier of the evaluation run that produced it.

Confinement is physical rather than conventional. A pure artifact's WebAssembly module has zero imports; the loader refuses any module whose imports exceed its manifest's declared capabilities; a tool-calling artifact imports exactly one host function, through which every tool invocation is named, auditable, and budget-limited. The sandbox, not a policy file, is what makes an artifact unable to exceed its declaration.

Guards are first-class because their failure mode is silent. A wrong ``stay compiled'' decision serves a stale answer that looks healthy; \S\ref{sec:eval} measures exactly this at 48.9\% wrongness under loose calibration. The conformal construction makes the exposure a declared quantity: $\alpha$ prices how far beyond witnessed inputs the artifact may speak.

\section{\autobench: Measuring Compiled Cognition}
\label{sec:eval}

Capability benchmarks ask how smart the model is. \autobench{} asks whether the \emph{system} ever pays for the same thought twice, and whether it knows what it does not know. The protocol declares four headline measurements: H1, the ratchet curve (cost per task as a function of experience); H2, the determinism census; H3, parity-gated compression; H4, calibrated ignorance. We report them in the order census, compression, ratchet, calibration, because each rests on the one before it. The protocol was frozen before execution (corpora, shift schedule, thresholds, anti-gaming rules) and is in the repository: every input is recorded at least twice (the witness rule), contracts with their thresholds and judge model are committed before any compile and never retuned after a verdict, example values may come only from recorded reality, the baseline and the compiled run consume the same frozen inputs, refusals are results, and every number carries an evaluation-run id, a spend-ledger session, or a committed CSV. Five protocol deviations occurred and are logged in the results document; none touch decision logic.

\paragraph{Setup.} Reference model: \texttt{gpt-5.4-mini} (provider snapshot \texttt{2026-03-17}), default temperature, pinned prices. Host: Windows 11, release builds; distillation uses scikit-learn decision trees. Six task families: F1 ticket triage (single call), F2 a three-call inbox pipeline with a tool, F3 typed field extraction, F4 policy routing with skewed classes, F5 free-text summarization under judged equivalence (F5 verifies the summarize span of F2's recorded store, so its frontier column is per call where F2's is per run), and F6 the novelty stream: the F1 triage task replayed as 300 items whose category set expands at three scheduled shifts, censused under the ratchet rather than under H2. Total benchmark spend on the compiler side: \$0.0621 against a pre-registered \$5 cap; a stranger with an API key reproduces every table for under a dollar.

\subsection{H2: the determinism census}

\begin{table}[t]
\centering
\caption{The determinism census. Witnessed $=$ observed $\geq 2$ times; a span is deterministic iff all witnesses agree byte-for-byte and none errored. Fractions are of witnessed spans only.}
\label{tab:census}
\small
\begin{tabular}{@{}lrrr@{}}
\toprule
Family & Spans & Witnessed & Deterministic \\
\midrule
F1 ticket triage & 80 & 80 & \textbf{100.0\%} \\
F2 inbox pipeline (all spans) & 320 & 320 & 77.5\% \\
\quad classify / priority / summarize / tool & & & 100.0 / 92.5 / 17.5 / 100.0\% \\
F3 field extraction & 80 & 80 & \textbf{100.0\%} \\
F4 policy routing & 80 & 80 & \textbf{100.0\%} \\
\midrule
Pooled & 560 & 560 & \textbf{87.1\%} \\
\bottomrule
\end{tabular}
\end{table}

This measurement could have killed the thesis: a low witnessed-deterministic fraction would have left nothing to compile. Table~\ref{tab:census} is the thesis, measured: 87.1\% of 560 real frontier-agent spans are witnessed-deterministic at default temperature. The texture matters more than the pooled number. Classification, extraction, and routing are fully deterministic; the free-text summarization span is 17.5\%; the pipeline's priority span sits at 92.5\%, deterministic enough to feel compilable and divergent enough to break an exact contract, which is precisely what happens below.

\subsection{H3: parity-gated compression, including the refusals}

\begin{table}[t]
\centering
\caption{Compression at declared parity. Frontier numbers are recorded per-call means (cost) and medians (latency); F2 is per 3-call run. Compiled latency is one-shot process wall-clock including spawn and module compile; resident floors are 290\,$\mu$s (stdio), 54.1\,$\mu$s (Python in-process), 18.2\,$\mu$s (Node in-process). Refusals are quoted outcomes, not omissions.}
\label{tab:compression}
\scriptsize
\setlength{\tabcolsep}{2.5pt}
\begin{tabular}{@{}llll@{}}
\toprule
Family & Frontier & Compiled & Gate outcome \\
\midrule
F1 triage & 55.6\,$\mu$\$ / 904\,ms & \$0 / 65.7\,ms & \textbf{PASS} 40/40 $\geq$ 950\textperthousand{} (both guard wires) \\
F4 routing (weighted) & 112.0\,$\mu$\$ / 1037\,ms & \$0 / 65.7\,ms & \textbf{PASS} 40/40 $\geq$ 900\textperthousand \\
F4 routing (holdout 0.25) & \multicolumn{1}{c}{--} & \$0 & \textbf{PASS} 36/40 $=$ 90.0\% at the exact threshold \\
F2 classify & \multirow{2}{*}{192.5\,$\mu$\$ / 2881\,ms$^{\dagger}$} & \$0 & \textbf{PASS} 40/40 (exact) \\
F2 priority &  & \multicolumn{1}{c}{--} & \textbf{FAIL}: 3 divergent-witness groups (exact contract) \\
F3 extraction & 255.8\,$\mu$\$ / 1171\,ms & \multicolumn{1}{c}{--} & \textbf{refused at all three rungs} (see text) \\
F5 summarize (judged, $n{=}40$) & 90.9\,$\mu$\$ / 1042\,ms & \multicolumn{1}{c}{--} & \textbf{FAIL} 17/40 $=$ 42.5\% $<$ 800\textperthousand{} (23 judge calls; 33 across F5) \\
\bottomrule
\multicolumn{4}{@{}l}{\scriptsize $^{\dagger}$per full 3-call run.}
\end{tabular}
\end{table}

Table~\ref{tab:compression} shows both sides. Where behavior is deterministic and expressible, compilation is a three-orders-of-magnitude cost cut at verified parity, with the F4 generalization run passing at exactly its declared 90.0\% threshold, the kind of razor-edge number a benchmark should be able to produce. Where it is not, the gate says so precisely. F3 is the sharpest refusal: a 100.0\%-deterministic extraction behavior that no current rung can compile, because its outputs are JSON objects; enumerative search has no object constructor, the LLM proposer's constant-output cheat was executed against a counterexample and rejected, and the tree trainer refuses non-string labels. F2's priority span shows why statistical acceptance exists: 92.5\% determinism is fatal to an exact contract. F5 shows the honest scale limit of memorization: at 40 tickets the free-text residue fails its declared judged threshold (a 20-ticket run passes at 85\%), so generative spans stay tier-0 until a generative rung exists.

\subsection{H1: the ratchet curve}

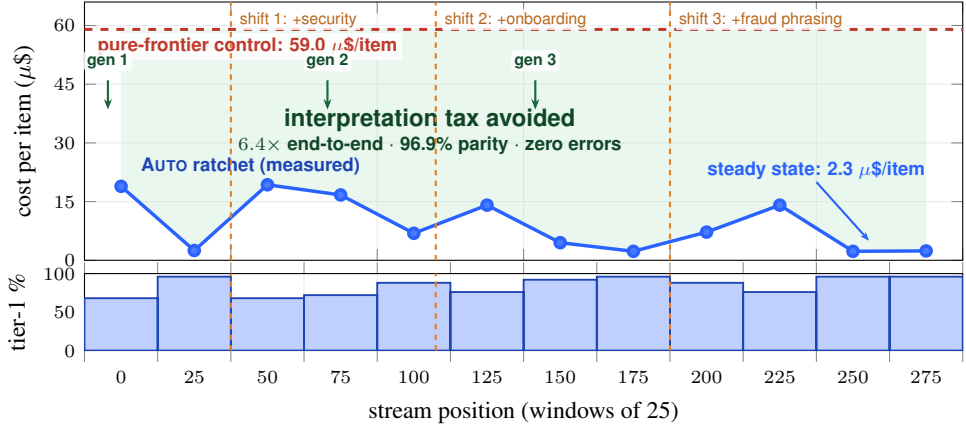
\begin{figure}[t]
\centering
\begin{tikzpicture}
\begin{groupplot}[
  group style={group size=1 by 2, x descriptions at=edge bottom, vertical sep=5pt},
  width=13.2cm,
  xmin=0, xmax=300,
  xtick={0,50,100,150,200,250,300},
  grid=major, grid style={black!8},
  tick label style={font=\scriptsize},
  label style={font=\footnotesize},
]
\nextgroupplot[
  height=5.0cm, ymin=0, ymax=66, ytick={0,15,30,45,60},
  ylabel={cost per item ($\mu$\$)},
  every axis plot/.append style={line width=1.3pt},
  clip=false,
]
\addplot[failred, dashed, line width=1.15pt, name path=A] coordinates {(0,59.0) (300,59.0)};
\node[font=\scriptsize\bfseries\sffamily, text=failred, anchor=north west, fill=white, fill opacity=0.85, text opacity=1, inner sep=1.5pt]
  at (axis cs:3,58.2) {pure-frontier control: 59.0 $\mu$\$/item};
\addplot[recblue, mark=*, mark size=1.9pt, name path=B, mark options={fill=recblue!85}] coordinates {
  (12.5,18.9) (37.5,2.5) (62.5,19.3) (87.5,16.7) (112.5,6.9) (137.5,14.1)
  (162.5,4.5) (187.5,2.3) (212.5,7.2) (237.5,14.1) (262.5,2.3) (287.5,2.4)
};
\addplot[runnergreen!35, fill opacity=0.28] fill between[of=A and B, soft clip={domain=12.5:287.5}];
\node[font=\footnotesize\bfseries\sffamily, text=runnergreen!40!black, align=center]
  at (axis cs:118,33) {interpretation tax avoided\\[-1pt]{\scriptsize $6.4\times$ end-to-end $\cdot$ 96.9\% parity $\cdot$ zero errors}};
\node[font=\scriptsize\bfseries\sffamily, text=recblue, anchor=south, inner sep=1pt] (ss)
  at (axis cs:250,20) {steady state: 2.3 $\mu$\$/item};
\draw[recblue, thick, -{Stealth[length=3.5pt]}] (ss.south) -- (axis cs:268,5);
\node[font=\scriptsize\bfseries\sffamily, text=recblue!70!black, anchor=west]
  at (axis cs:16,24) {\auto{} ratchet (measured)};
\draw[gateorange, thick, dash pattern=on 2pt off 2pt] (axis cs:50,0) -- (axis cs:50,66);
\draw[gateorange, thick, dash pattern=on 2pt off 2pt] (axis cs:120,0) -- (axis cs:120,66);
\draw[gateorange, thick, dash pattern=on 2pt off 2pt] (axis cs:200,0) -- (axis cs:200,66);
\node[font=\tiny\sffamily, text=gateorange!80!black, anchor=west, fill=white, fill opacity=0.8, text opacity=1, inner sep=1pt] at (axis cs:52,61.5) {shift 1: +security};
\node[font=\tiny\sffamily, text=gateorange!80!black, anchor=west, fill=white, fill opacity=0.8, text opacity=1, inner sep=1pt] at (axis cs:122,61.5) {shift 2: +onboarding};
\node[font=\tiny\sffamily, text=gateorange!80!black, anchor=west, fill=white, fill opacity=0.8, text opacity=1, inner sep=1pt] at (axis cs:202,61.5) {shift 3: +fraud phrasing};
\draw[-{Stealth[length=4pt]}, runnergreen!60!black, thick] (axis cs:8,46) -- (axis cs:8,38.5);
\node[font=\tiny\bfseries\sffamily, text=runnergreen!45!black, fill=white, fill opacity=0.85, text opacity=1,
      inner sep=1.2pt, rounded corners=1.5pt] at (axis cs:8,51) {gen 1};
\draw[-{Stealth[length=4pt]}, runnergreen!60!black, thick] (axis cs:83,46) -- (axis cs:83,38.5);
\node[font=\tiny\bfseries\sffamily, text=runnergreen!45!black, fill=white, fill opacity=0.85, text opacity=1,
      inner sep=1.2pt, rounded corners=1.5pt] at (axis cs:83,51) {gen 2};
\draw[-{Stealth[length=4pt]}, runnergreen!60!black, thick] (axis cs:154,46) -- (axis cs:154,38.5);
\node[font=\tiny\bfseries\sffamily, text=runnergreen!45!black, fill=white, fill opacity=0.85, text opacity=1,
      inner sep=1.2pt, rounded corners=1.5pt] at (axis cs:154,51) {gen 3};
\nextgroupplot[
  height=2.6cm, ymin=0, ymax=100, ytick={0,50,100},
  ylabel={tier-1 \%},
  xlabel={stream position (windows of 25)},
  ybar interval, bar shift=0pt,
]
\addplot[fill=recblue!30, draw=recblue!70!black, line width=0.7pt] coordinates {
  (0,68) (25,96) (50,68) (75,72) (100,88) (125,76)
  (150,92) (175,96) (200,88) (225,76) (250,96) (275,96) (300,96)
};
\draw[gateorange, thick, dash pattern=on 2pt off 2pt] (axis cs:50,0) -- (axis cs:50,100);
\draw[gateorange, thick, dash pattern=on 2pt off 2pt] (axis cs:120,0) -- (axis cs:120,100);
\draw[gateorange, thick, dash pattern=on 2pt off 2pt] (axis cs:200,0) -- (axis cs:200,100);
\end{groupplot}
\end{tikzpicture}
\caption{The ratchet curve, measured on a 300-item stream with three scheduled distribution shifts (orange). \emph{Top:} marginal cost per item collapses from 18.9 to 2.3\,$\mu$\$ across three gate-passed recompile generations (green; the driver recompiles after every $K{=}8$ new distinct ingested inputs, so generation 1 is the bootstrap compile at position 8, and generations 2 and 3 land one compile cycle behind shifts 1 and 2). Shift 3's phrasing novelty is absorbed by the lexical guard rather than a recompile, the calibration exposure \S\ref{sec:eval} quantifies. The shaded region is the money not paid to the interpreter (stream total 2{,}775 vs.\ a 17{,}692\,$\mu$\$ control, $6.4\times$; the control is arithmetic: each position priced at its measured tier-0 cost, 218 of 300 estimated from the recorded mean, never re-executed; 253/300 items compiled; zero errors; 96.9\% parity on witnessed inputs). \emph{Bottom:} the fraction of each window served by the compiled tier. Per-position CSV and evaluation-run ids ship with the repository.}
\label{fig:ratchet}
\end{figure}

Figure~\ref{fig:ratchet} is the paper's central measurement. The system starts uncompiled, pays the interpreter while bootstrapping, and compiles generation 1 at position 8. Shifts 1 and 2 inject unseen categories; guards trip, items deopt to the reference agent, and generations 2 and 3 land one compile cycle later. Shift 3 plays out differently and honestly: its fraud-phrased tickets are lexical cousins of witnessed billing text, the guard admits most of them, fewer than $K{=}8$ new distinct inputs are ever ingested, and no fourth generation fires; the window recovers through guard admission, not recompilation, which is precisely the exposure H4 prices. Steady-state windows cost 2.3 to 2.5\,$\mu$\$/item against a 59.0\,$\mu$\$ interpreted control, with every recompile passing the same differential gate that governs offline compilation. Nothing is figured out twice: the stream contains 56 distinct ticket texts across 300 positions, and repetition is exactly what the ratchet monetizes.

\begin{table}[t]
\centering
\caption{Three operating points of the same frozen stream. Loose calibration is cheap and silently wrong; a generic deopt reference stalls the ratchet at the gate; the agent-faithful configuration closes the loop. ``Silently wrong'' compares tier-1 answers to stream design labels.}
\label{tab:legs}
\small
\setlength{\tabcolsep}{4pt}
\begin{tabular}{@{}lllrrr@{}}
\toprule
Leg & Guard $\alpha$ & Deopt reference & Total cost & Recompiles & Tier-1 silently wrong \\
\midrule
A & 0.100 & generic frontier & 7{,}770\,$\mu$\$ & 1 & \textbf{48.9\%} (133/272) \\
B & 0.400 & generic frontier & 33{,}387\,$\mu$\$ & 1 (+3 \textbf{gate-refused}) & 28.4\% (55/194) \\
C & 0.400 & the agent itself & \textbf{2{,}775\,$\mu$\$} & \textbf{3 (all passed)} & see text \\
\bottomrule
\end{tabular}
\end{table}

Table~\ref{tab:legs} is the finding we consider most important. Leg A looks operationally perfect, 272 of 300 items served compiled at trivial cost, while 48.9\% of those answers disagree with the stream's design labels: the loose guard admits post-shift novelty and the stale artifact mislabels it (\texttt{security}\,$\to$\,\texttt{bug} 32 times). Leg B tightens the guard and halves the damage, but its recompiles are \emph{refused by the gate}: the generic tier-0 interpreter answers outside the contract's vocabulary, and the system correctly declines to compile a non-conformant reference, stalling the ratchet honestly. Only leg C, where deopts re-record through the agent's own prompt, closes the loop. Two separations follow. First, calibration and reference fidelity decide whether cheap stays correct; model capability does not. Second, parity is not ground truth: leg C's tier-1 answers match the reference agent on 96.9\% of witnessed inputs (witnessed here meaning previously recorded by bootstrap or deopt) while 53.8\% differ from the six-label stream design, because the reference agent itself is a three-label agent; the compiler faithfully preserves the agent you have, including its blindness.

\subsection{H4: calibrated ignorance}

On frozen probe sets (five held-out in-distribution and five disjoint-vocabulary out-of-distribution probes for each family that emitted an artifact, F1 and F4), the F4 artifact is perfectly calibrated: zero false-abstains (in-distribution probes needlessly refused) and zero false-proceeds (out-of-distribution probes answered compiled, the silent failure). F1 shows the lexical noise floor: 2/5 OOD probes admitted on the trigram-Jaccard wire and 3/5 on the hashed-cosine wire at maximal quantile, consistent across the static probes and the deployed stream. The guard geometry is lexical, so lexical cousins get in; the leg-C residue quantifies the exposure precisely, 125 of 253 tier-1 answers rode on never-witnessed inputs the guard admitted. Semantic embedding guards are the recorded upgrade, and $\alpha$ is the declared price of the boundary in the meantime.

\section{Limitations}
\label{sec:limitations}

The corpora are designed, not production traffic: 20 to 40 recorded inputs per family (the stream holds 56 distinct texts), one reference model, and repetition rates chosen to exercise the ratchet; the protocol is built to be re-run unchanged on real traffic, and that re-run is the experiment that matters. The compile rungs are v0: outputs must be strings or DSL-expressible values, which is exactly why F3 refuses, and the generative residue (17.5\%-deterministic summarization) does not compile at 40-ticket scale under its judged threshold. Guards are lexical; a paraphrase with disjoint vocabulary trips, and a lexical cousin with different meaning may be admitted, an exposure we measure rather than solve. Judged verification inherits the judge's opinions and is reproducible only up to the judge model named in the manifest. Compiled artifacts memorize witnessed distributions unless a holdout demonstrates generalization; the gate makes this explicit rather than making it false.

\section{Conclusion}

\auto{} treats agent cognition the way compilers treat interpreted code: record it, prove what is symbolic, distill what is not, verify everything against a contract, and emit a confined binary that a tiered runtime executes for microdollars, deopting and recompiling when the world drifts. On \autobench, 87.1\% of 560 recorded frontier-agent spans are witnessed-deterministic, compiled families run at three orders of magnitude below interpretation at declared parity, and the closed loop answers three distribution shifts with three gate-passed recompilations and a measured calibration exposure, at 2\,$\mu$\$ per item steady state. The same harness quantifies the failure modes that make naive compilation dangerous. The toolchain, benchmark protocol, evidence, and spend ledgers are at \url{https://github.com/RightNow-AI/auto}.

\end{document}